\documentclass{article}
\usepackage[square,sort,comma,numbers,compress]{natbib}
\usepackage[final]{neurips_2020}

\usepackage[utf8]{inputenc} 
\usepackage[T1]{fontenc}    
\usepackage{hyperref}       
\usepackage{url}            
\usepackage{booktabs}       
\usepackage{amsfonts}       
\usepackage{nicefrac}       
\usepackage{microtype}      
\usepackage{lipsum}
\usepackage{graphicx}
\graphicspath{{figures/}}
\usepackage{subcaption}
\usepackage{floatrow}
\usepackage{siunitx}
\usepackage{xspace}
\usepackage[export]{adjustbox}

\makeatletter
\DeclareRobustCommand\onedot{\futurelet\@let@token\@onedot}
\def\@onedot{\ifx\@let@token.\else.\null\fi\xspace}

\def\ie{\emph{i.e}\onedot}

\makeatother

\title{%
	Generating Novel Glyph without Human Data
	by Learning to Communicate
}

\author{%
  Seung-won Park\thanks{\url{http://swpark.me}} \\
  Seoul National University, MINDs Lab Inc. \\
  \texttt{yyyyy@snu.ac.kr} \\
}

\begin{document}
\maketitle
\begin{abstract}
	In this paper, we present \textit{Neural Glyph},
	a system that generates novel glyph without any training data.
	The generator and the classifier are trained to communicate
	via visual symbols as a medium,
	which enforces the generator to come up with a set of distinctive symbols.
	Our method results in glyphs that resemble the human-made glyphs,
	which may imply that the visual appearances of existing glyphs
	can be attributed to constraints of communication via writing.
	Important tricks that enable this framework are described
	and the code is made available.
\end{abstract}

\section{Introduction}

The glyph is a visual representation of characters for writing,
which is a medium of human communication along with speech.
Throughout human history, hundreds of glyphs were created,
and some of them are still in use today.
Despite many of the glyphs were invented independently,
some of the glyphs are known to resemble each other
\cite{changizi2006structures, morin2018spontaneous}.
But then, if artificial intelligence is given a task
to generate the glyph for visual communication
without any training data derived from human's writing,
will it resemble the existing glyphs?
What will such `alien language' look like?

In this paper,
we propose \textit{Neural Glyph}, a system that can generate novel glyph
by training neural networks
to communicate via visual symbols.
We find that such `alien language' looks visually similar to existing glyphs,
even though any data from existing glyphs are used.
We also propose a method for controlling
the amount of variation of visual appearance within each symbol.

\section{System Architecture}

\begin{figure}[t]
	\centering
	\includegraphics[width=\textwidth]{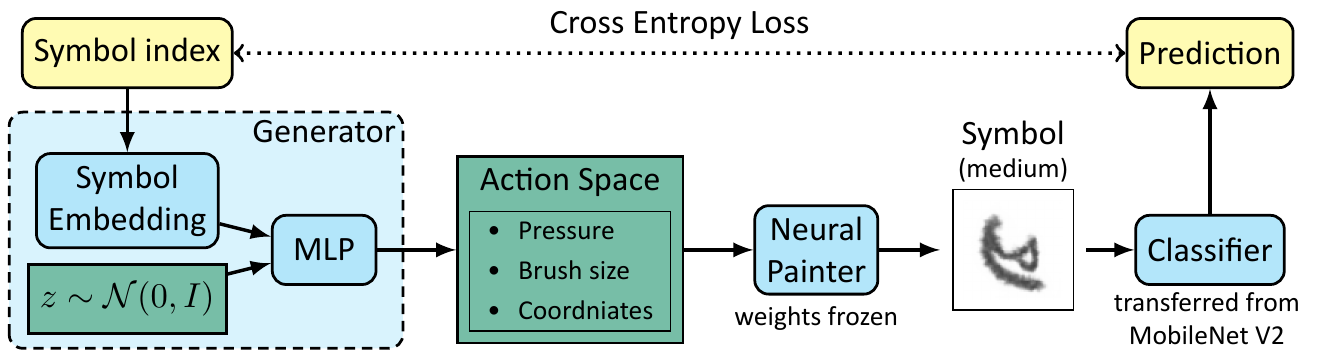}
	\caption{%
		System architecture of Neural Glyph.
		Yellow, blue, green colors of the boxes denote
		input/output, trainable modules, and intermediate representations,
		respectively.
	}
	\label{fig:arch}
\end{figure}

Our system aims to generate a novel set of symbols
by jointly training a generator and classifier
to transfer a message via visual symbols as a medium (channel),
following the Shannon-Weaver model of communication \cite{shannon1948mathematical}.
The whole system is trained with only classification loss in an end-to-end manner,
as shown in figure \ref{fig:arch}.
The generator encodes the given message into a sequence of action spaces
that define the brushstroke for a symbol.
Then, the sequence of action space is rendered into a visual symbol
with a pre-trained neural painter \cite{neuralpainter}.
Finally, the classifier predicts the original message
that the generator aimed to deliver via the symbol.

\paragraph{Generator.}
The generator aims to synthesize the sequence of action space
that defines the visual appearance of the given index (message) of symbol.
To generate the glyph consisted of $ N $ symbols,
the embedding lookup table of size $ N $ is constructed
to provide representation for each distinctive symbols.
Then, a two-layer MLP transforms the representation into
the sequence of action space.
Considering the fact that handwriting produced by humans is different
for every individual and time,
we try to simulate the stochastic writing behavior
by injecting random noise to the MLP.

\paragraph{Neural Painter.}
To propagate the gradients from prediction error to the whole system,
it is critical for the renderer to be differentiable.
In our system, we adopt a GAN-based neural painter \cite{neuralpainter},
which was trained with a synthetic brushstroke dataset
consisted of B\'{e}zier curves.
We use the PyTorch implementation and the pre-trained weights
produced by the author of the neural painter \cite{neuralpainter}.
While training our system, the weights of neural painter are frozen.
To make the generator focus on generating a set of symbols that have diverse shapes,
we fix the brushstroke to have constant thickness and color (black)
by setting the corresponding action parameters constant.

\paragraph{Classifier.}
The CNN-based classifier is trained
to decipher the original message from the rendered symbol.
The classifier is built with a pre-trained MobileNet V2 \cite{mobilenetv2},
where the last fully-connected layer is replaced with
a randomly initialized projection layer to produce logits for symbol classification.
We empirically observed that the system fails to generate the distinctive symbols
when the classifier is trained from scratch
or the weights from MobileNet V2 are frozen.

\section{Experiments}

\paragraph{Results.}

\begin{figure}[t]
	\floatbox[{\capbeside\thisfloatsetup{%
	capbesideposition={right,bottom},
	capbesidewidth=0.48\textwidth}}]{figure}[\FBwidth]
	{\includegraphics[width=0.45\textwidth]{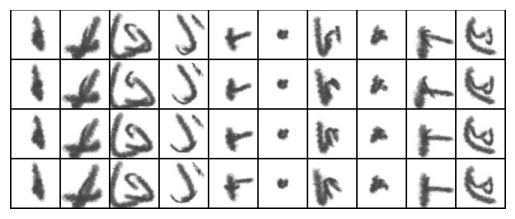}}
	{\caption{%
		An example of generated glyph consisted of $ N=10 $ symbols,
		generated with temperature 1.0.
		For each symbols, four different examples are shown to illustrate
		the stochasticity of result for each symbol.
	}\label{fig:example}}
\end{figure}

The resulting glyph is different for each run
since it is possible to generate
a limited number of distinctive symbols in various ways.
One of the example results is shown in figure \ref{fig:example},
and more results are shown in figure \ref{fig:more_examples}.
We may observe that many of the results resemble
the symbols from the existing glyphs;
we encourage the readers to compare the results with
glyphs shown in the Omniglot encyclopedia \cite{omniglot},
since most of them are not in use for nowadays.

\paragraph{Effect of temperature.}
While running the system after training,
the amount of variation of visual appearance within each symbol
can be adjusted by controlling the 
temperature (magnitude) of the noise injected through the generator.
The results are shown in figure \ref{fig:temperatures}.

\section{Future Works}
We have shown the preliminary results on
generating novel glyphs without human data by learning to communicate.
Our results are encouraging and numerous future works will be possible.
First, constructing an action space
that better represents the human brushstroke
may lead to results that better resembles the human-made glyphs,
where only three B\'{e}zier curves are used in this work.
Furthermore, exploring the results 
with other types of medium for communication will be interesting;
this might be possible by utilizing their corresponding neural renderers
with appropriate constraints applied,
such as an audio signal (DDSP \cite{ddsp}) or 3D object (Softras \cite{liu2019soft}).

\clearpage 

\section*{Broader Impact}
The emergence of the novel glyph from communication between neural networks
may inspire artists, and benefit 
linguists or cognitive scientists on studying the underlying pattern of
existing glyphs.
However, we believe the amount of impact will be limited
since the real-world application of our work does not seem to exist.

\section*{Acknowledgments}
The author would like to thank
Bokyung Son (Kakao Enterprise),
Donghwan Jang (Seoul National University),
and the members of Deepest AI for their insightful discussions.
Deepest AI is the Seoul National University Deep Learning Society.
\footnote{\url{http://deepest.ai/}}

\bibliographystyle{plainnat}
\bibliography{swpark_neurips2020}

\appendix


\section{Related Works}

Previous works on generating novel visual language
that works with a communication objective include
Cooperative Communication Networks \cite{murdock_2020},
Dimension of Dialogue \cite{simon_2019},
and GlyphNet \cite{trenaman_2020}.
However, our work is the first to generate
a visual language based on glyphs, which is directly interpretable by human vision.

\section{Reproducibility}
To enhance the reproducibility of our work,
we provide some detailed information about our system.

\paragraph{Code availability.}

A jupyter notebook script for training 
is available at the following URL
\footnote{\url{https://colab.research.google.com/drive/1NDEdM7PjcS2ohKP39UnsX02hg_EyOpYX?usp=sharing}}.

\paragraph{Hyperparameters.}

\begin{itemize}
	\item Number of symbol classes to form a glyph ($ N $): 10
	\item Number of strokes (length of the sequence of action space): 3
	\item Batch size: 16
	\item Optimizer: Adam (learning rate \num{1e-3})
	\item Number of training steps: 10k (takes 5 minutes on Google Colab P100 GPU)
	\item Symbol embedding dimension: 16
	\item Generator noise vector dimension: 16
	\item Generator MLP dimension: 32
	\item Action space parameters that are controlled by the generator:
	\texttt{pressure}, \texttt{control\_x}, \texttt{control\_y},
	\texttt{end\_x}, \texttt{end\_y}, \texttt{start\_x}, \texttt{start\_y},
	\texttt{entry\_pressure}
	\item Action space parameters that are fixed:
	\texttt{size} (brush size, set to 0),
	\texttt{color\_r}, \texttt{color\_g}, \texttt{color\_b} (colors, set to -1)
\end{itemize}

\paragraph{Miscellaneous notes.}

\begin{itemize}
	\item Throughout the training iteration,
	the generated set of symbols do not tend to converge into constant visual appearance,
	and the loss with respect to the number of iteration spikes.
	The training should be stopped when
	the validation loss is reasonably low or the user gets satisfactory results.
	Failing to stop at the right time may lead to redundant symbols.
	\item Implementing our system does not require the data loader,
	since the input/target batches are constructed with random symbol indices.
\end{itemize}


\section{Accuracy of the Classifier}

Representing and classifying the number of distinctive symbols within limited action space
will be more challenging if the number of categories (\ie, $ N $) is larger.
To validate such hypothesis, we measure the top-1 accuracy of the classifier
with respect to $ N $.
For each $ N $, the experiment is repeated for 7 times with different random seeds.
The accuracy is measured with 10k randomly generated samples,
while the number of training steps is fixed to the default setting (10k).

The mean/variance of accuracy for each $ N $ are shown in table \ref{tab:n_acc}.
It is clearly visible that the system finds it hard to
correctly represent and classify the symbols when $ N $ is larger.

\begin{table}[h]
	\centering
		\caption{%
			Top-1 accuracy of the classifier
			with respect to the number of symbols ($ N $).
		}
		\label{tab:n_acc}
		\begin{tabular}{c|c}
		$ N $ & Top-1 Accuracy (\%) \\
		\hline
		\phantom{00}4 & $ 99.80 \pm 0.03 $ \\
		\phantom{00}8 & $ 99.24 \pm 0.70 $ \\
		\phantom{0}16 & $ 98.47 \pm 1.35 $ \\
		\phantom{0}32 & $ 98.52 \pm 0.48 $ \\
		\phantom{0}64 & $ 96.54 \pm 0.73 $ \\
		128 & $ 94.43 \pm 3.08 $ \\
		256 & $ 90.02 \pm 5.39 $ \\
		512 & $ 87.84 \pm 18.47 $ \\
	\end{tabular}
\end{table}


\clearpage

\section{More Examples of Generated Glyph}

\begin{figure}[h]
\centering

\begin{tabular}{cc}
& \parbox[c]{1.5cm}{\centering Top-1 Accuracy}  \\
\includegraphics[width=0.8\textwidth, valign=m]{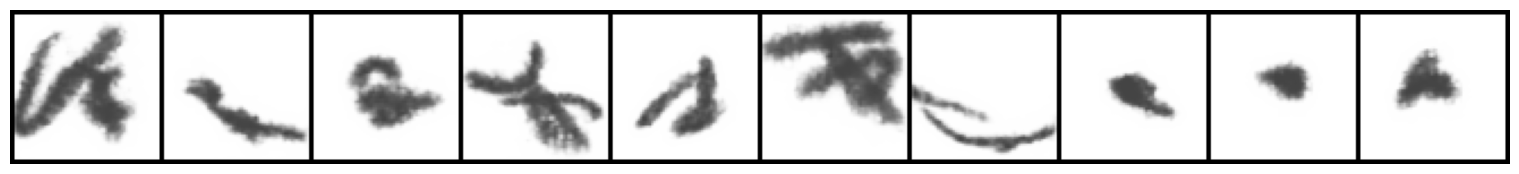}
& \parbox[c]{1.5cm}{\centering 99.54\%} \\
\includegraphics[width=0.8\textwidth, valign=m]{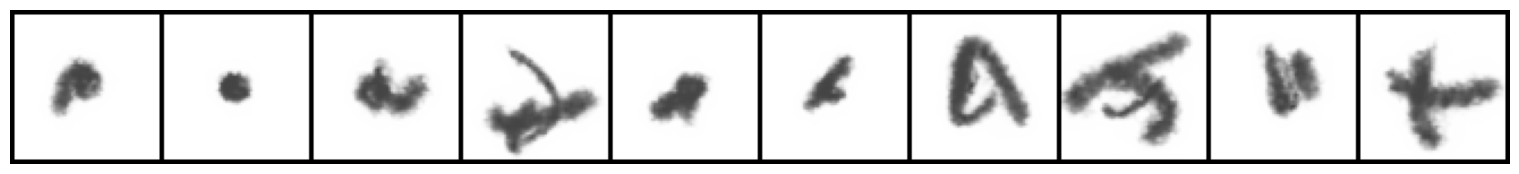}
& \parbox[c]{1.5cm}{\centering 98.80\%} \\
\includegraphics[width=0.8\textwidth, valign=m]{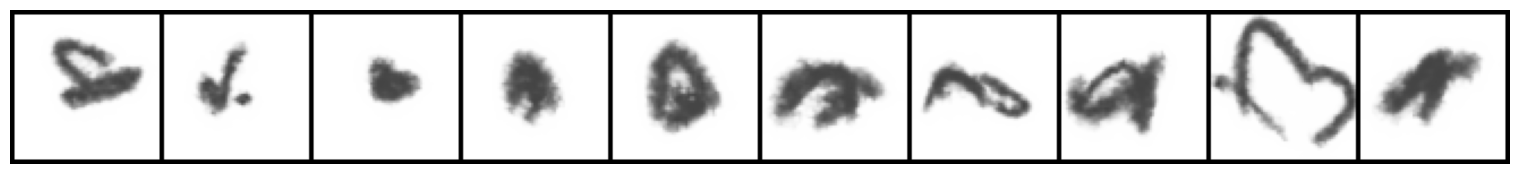}
& \parbox[c]{1.5cm}{\centering 98.58\%} \\
\includegraphics[width=0.8\textwidth, valign=m]{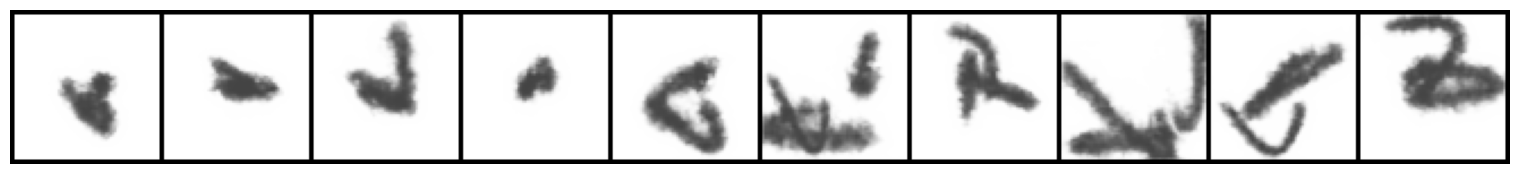}
& \parbox[c]{1.5cm}{\centering 99.82\%} \\
\includegraphics[width=0.8\textwidth, valign=m]{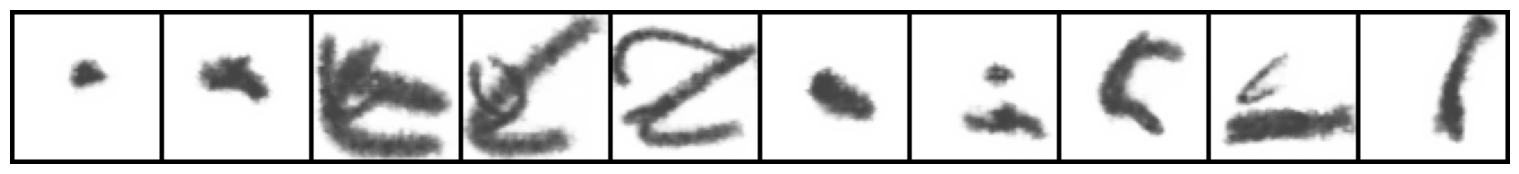}
& \parbox[c]{1.5cm}{\centering 99.68\%} \\
\includegraphics[width=0.8\textwidth, valign=m]{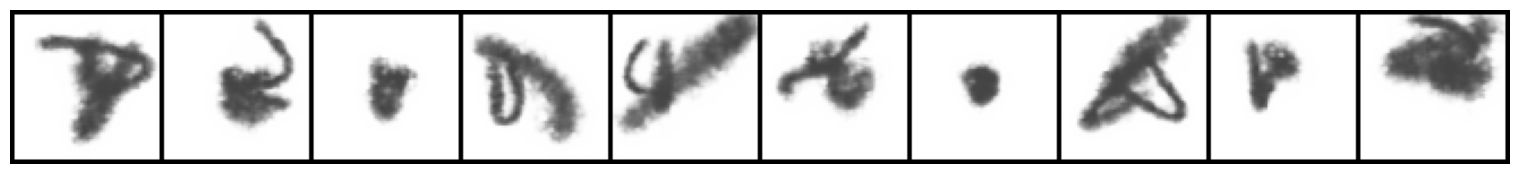}
& \parbox[c]{1.5cm}{\centering 99.75\%} \\
\includegraphics[width=0.8\textwidth, valign=m]{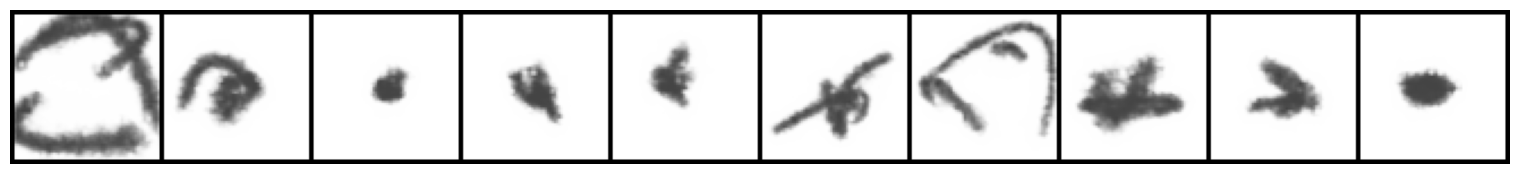}
& \parbox[c]{1.5cm}{\centering 99.24\%} \\
\includegraphics[width=0.8\textwidth, valign=m]{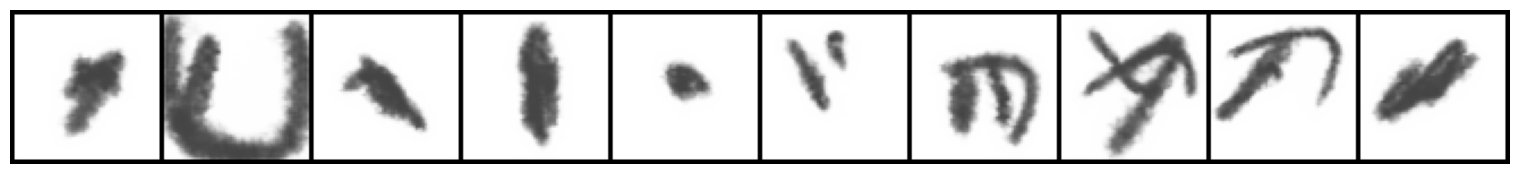}
& \parbox[c]{1.5cm}{\centering 99.58\%} \\
\includegraphics[width=0.8\textwidth, valign=m]{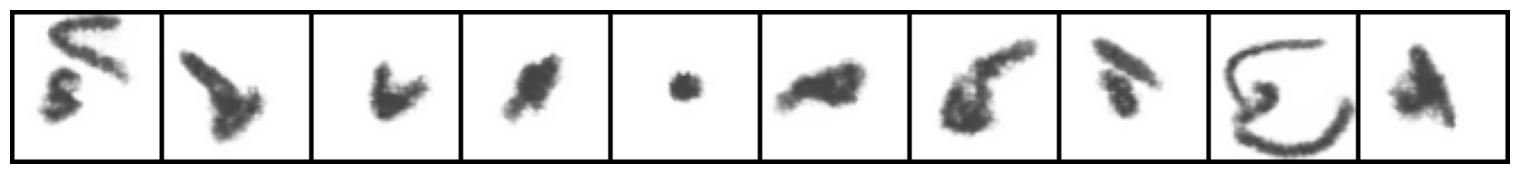}
& \parbox[c]{1.5cm}{\centering 99.54\%} \\
\includegraphics[width=0.8\textwidth, valign=m]{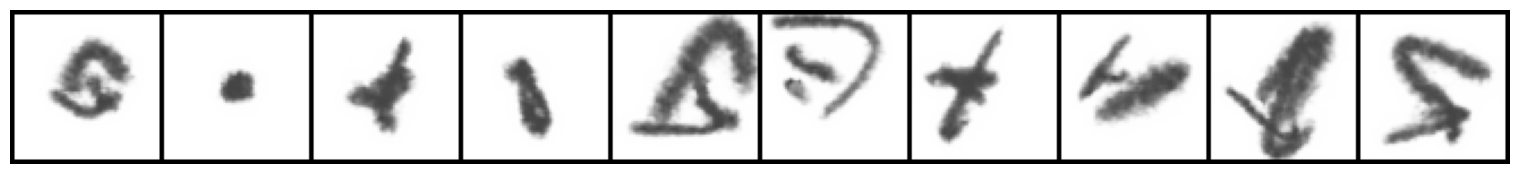}
& \parbox[c]{1.5cm}{\centering 99.15\%} \\
\end{tabular}

\caption{%
	Uncurated examples of generated glyph sampled with temperature $ T=1.0 $,
	showing results of every independent run in each row.
	It can be observed that the system prefers to leave one of the symbols
	as the centered dot,
	which is the result of randomly initialized weights of the generator.
	The top-1 accuracy of the classifier is measured
	with 10k randomly generated samples,
	and shown in the right side of the corresponding examples from each experiment.
}
\label{fig:more_examples}
\end{figure}

\clearpage

\setlength{\fboxrule}{2.5pt}

\begin{figure}[h]
\begin{tabular}{cp{0.7\textwidth}}

\begin{tabular}{c}
\fbox{\includegraphics[width=0.15\linewidth]{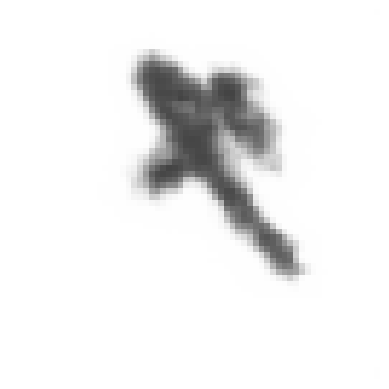}}
\end{tabular}
&
\begin{tabular}{c}
\includegraphics[width=\linewidth]{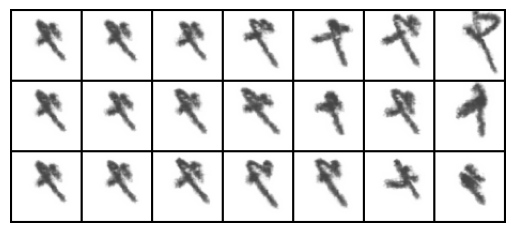}
\end{tabular}
\\
\begin{tabular}{c}
	\fbox{\includegraphics[width=0.15\linewidth]{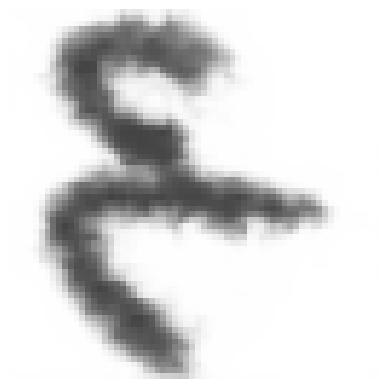}}
\end{tabular}
&
\begin{tabular}{c}
	\includegraphics[width=\linewidth]{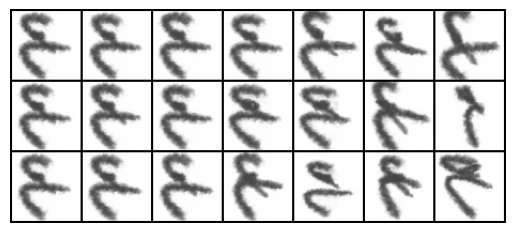}
\end{tabular}
\\
\begin{tabular}{c}
	\fbox{\includegraphics[width=0.15\linewidth]{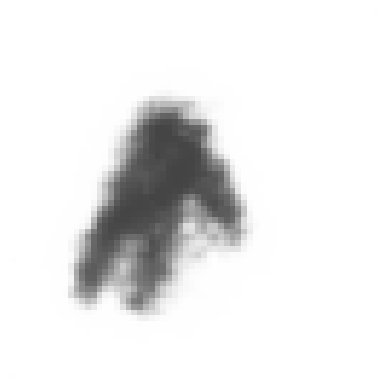}}
\end{tabular}
&
\begin{tabular}{c}
	\includegraphics[width=\linewidth]{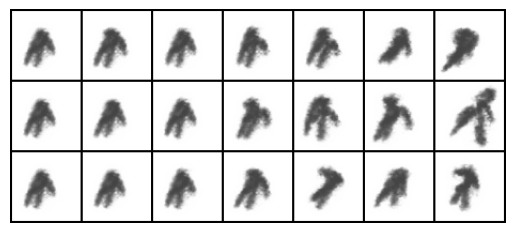}
\end{tabular}
\\
\begin{tabular}{c}
	\fbox{\includegraphics[width=0.15\linewidth]{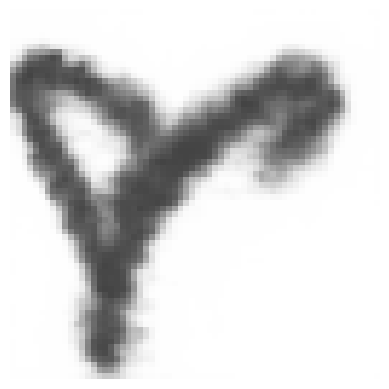}}
\end{tabular}
&
\begin{tabular}{c}
	\includegraphics[width=\linewidth]{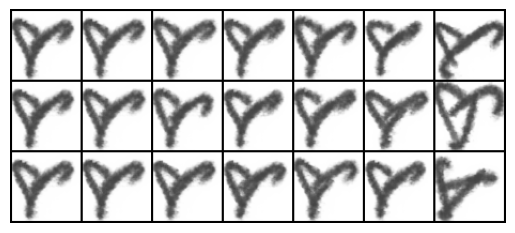}
\end{tabular}

\end{tabular}
\caption{Examples illustrating the amount of variation of visual appearance within each generated symbol, which is adjusted by controlling the temperature. For each row, a figure on the right side shows the samples generated with difference temperatures
($ T=0.0, 0.25, 0.5, 1.0, 1.5, 2.0, 4.0 $ for each column)
for the symbol shown in the left side.
Higher temperatures yield more stochastic results.
}
\label{fig:temperatures}
\end{figure}

\end{document}